\documentclass[conference]{IEEEtran}
\IEEEoverridecommandlockouts

\usepackage{cite}
\usepackage{amsmath,amssymb,amsfonts}
\usepackage{algorithmic}
\usepackage{graphicx}
\usepackage{textcomp}
\usepackage{adjustbox}
\usepackage{multirow}
\usepackage{booktabs}
\usepackage{booktabs}
\usepackage{adjustbox}
\usepackage{tabularx}
\usepackage{array}

\usepackage[caption=false,font=footnotesize]{subfig}

\usepackage{xcolor}
\def\BibTeX{{\rm B\kern-.05em{\sc i\kern-.025em b}\kern-.08em
    T\kern-.1667em\lower.7ex\hbox{E}\kern-.125emX}}

\title{Automated Prediction of Postoperative Pancreatic Fistula Using Preoperative Computed Tomography
}

\author{
\IEEEauthorblockN{
Ashok Choudhary\IEEEauthorrefmark{1},
Chris Varghese\IEEEauthorrefmark{1}\IEEEauthorrefmark{2},
Leo Y. Li-Han\IEEEauthorrefmark{1},
Frank G. Lee\IEEEauthorrefmark{1},
Ellen L. Larson\IEEEauthorrefmark{1}, \\
        Elizabeth B. Habermann\IEEEauthorrefmark{3}, 
        Cornelius A. Thiels\IEEEauthorrefmark{1},
Hojjat Salehinejad\IEEEauthorrefmark{3}\IEEEauthorrefmark{4}, Senior Member, IEEE
}
\IEEEauthorblockA{\IEEEauthorrefmark{1}Department of Surgery, Mayo Clinic, Rochester, MN, USA}
\IEEEauthorblockA{\IEEEauthorrefmark{2}Department of Surgery, University of Auckland, Auckland, NZ}
\IEEEauthorblockA{\IEEEauthorrefmark{3}Kern Center for the Science of Health Care Delivery, Mayo Clinic, Rochester, MN, USA}
\IEEEauthorblockA{\IEEEauthorrefmark{4}Department of Artificial Intelligence and Informatics, Mayo Clinic, Rochester, MN, USA}

\IEEEauthorblockA{
\begin{minipage}{\linewidth}\centering
\texttt{\{choudhary.ashok, varghese.chris, li-han.leo, lee.frank, larson.ellen1,}\\
\texttt{habermann.elizabeth, thiels.cornelius, salehinejad.hojjat\}@mayo.edu}
\end{minipage}
}
\thanks{This paper was accepted for presentation at the 2026 IEEE Engineering in Medicine and Biology Society (EMBC), Toronto, Canada, July 2026}
}

\begin{document}

\maketitle

\begin{abstract}
Postoperative pancreatic fistula (POPF) is a serious complication after pancreatic resection, increasing morbidity, hospital stay, and healthcare costs. We present an automatic, end-to-end deep learning pipeline—from pancreatic segmentation to classification—for preoperative POPF risk estimation and stratification using preoperative CT scans. A data set with auto-segmented pancreas volumes and surgical outcomes was used to evaluate multiple architectures, including a custom lightweight 3D CNN baseline (CNN3D), R(2+1)D ResNet-18, ResNet-MC3-18 models. Evaluation across multiple 3D architectures demonstrated promising predictive performance. This approach offers a clinically valuable tool and a methodological benchmark for pancreas-specific CT classification, supporting improved preoperative decision-making in pancreatic surgery.
\end{abstract}

\section{Introduction}
\label{sec:intro}

Postoperative pancreatic fistula (POPF) is a devastating complication following pancreatic resection. Despite improvements in surgical care, POPF remains a dominant driver of morbidity and mortality in pancreas surgery\cite{bassi20172016}\cite{theijse2025nationwide} .  The incidence of clinically relevant POPF ranges from about 10$\%$ to 40$\%$ of patients, depending on the surgical context, and this complication can lead to prolonged hospital stays, additional resource utilisation, and worse overall outcomes~\cite{bassi20172016,theijse2025nationwide,bassi2005postoperative,nahm2018postoperative,van2022incidence}. Preoperative identification of high-risk patients is therefore critical, enabling surgeons to tailor perioperative management and potentially mitigate the impact of POPF. Traditionally, risk stratification has relied on clinical factors and simple scoring systems. Here, we frame CT-based modeling as preoperative risk stratification: leveraging information available on CT to better characterize risk before surgery, rather than replacing intraoperative assessment. Features such as elevated body mass index (BMI), small pancreatic duct diameter, and soft gland texture are well-known predictors, frequently incorporated into bedside risk scores~\cite{nahm2018postoperative}\cite{gaujoux2010fatty}. 

\begin{figure*}[t!]
    \centering
    \makebox[\textwidth][c]{%
    \rotatebox[origin=c]{0}{%
        \includegraphics[
            width=0.85\textheight,          
            trim={0mm 0mm 0mm 0mm},     
            clip
        ]{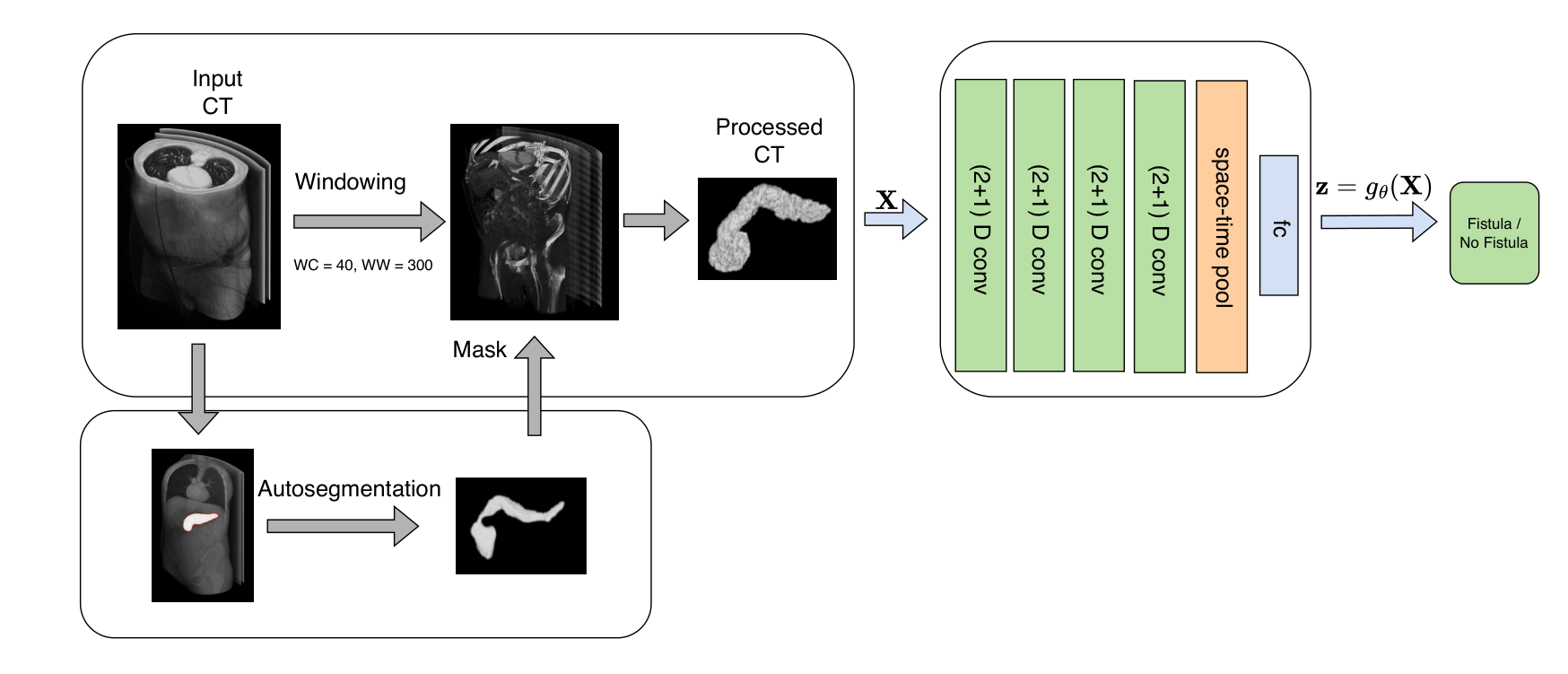}%
    }}%
    \caption{End-to-end POPF prediction from preoperative CT. We first apply Hounsfield-unit (HU) windowing to standardize intensities, then use a pancreas mask (MS/TS) to generate a tight pancreas-centric 3D ROI via masking and cropping/padding to a fixed size. The resulting volume is input to the classifier to estimate the probability of clinically relevant POPF.}
    \label{fig:model_pipeline}
\end{figure*}

While these scores are convenient and easy to use, their accuracy remains moderate, and they rely on often subjective and intraoperative data points. There is therefore a substantial potential to leverage the rich preoperative information available in radiology scans. Recent studies suggest that advanced imaging characteristics of the pancreas and surrounding anatomy can aid in the prognostication of POPF\cite{ingwersen2023radiomics}. Newly identified radiological factors such as the proportion of pancreatic fat or muscle (e.g., high visceral and subcutaneous fat indices) have been shown to correlate with fistula risk\cite{lee2024deep}, underscoring the  value of a imaging-based risk assessment. In principle, a detailed analysis of preoperative CT scans could reveal subtle textural features or anatomic variations that predispose to POPF features that may be difficult for human observers to quantify but are amenable to computational extraction. In particularl, known correlates with POPF such as intraoperatively palpated gland texture, could hypothetically be extrapolated based on imaging characteristics of the pancreas gland. This has led to growing interest in the application of radiomics and artificial intelligence approaches to analysis of high-dimensional image data to uncover complex patterns associated with postoperative complications. 

Deep learning (DL) techniques offer a powerful method to improve POPF risk prediction by automatically learning predictive features from imaging data. Unlike traditional statistical models, DL methods can capture non-linear relationships and interactions between multiple variables, potentially improving predictive performance. Several preliminary studies have explored machine learning ML and DL models using preoperative clinical parameters and CT images to predict POPF \cite{ingwersen2024radiomics}\cite{bhasker2023prediction}. However, these prior works were often limited by relatively small sample sizes or single-center cohorts, and most evaluated only a single algorithm or did not compare different modeling approaches \cite{ingwersen2024radiomics}. Consequently, there remains a need for a robust, systematic investigation of modern deep learning techniques for POPF prediction, using larger datasets and rigorous validation to establish real-world utility. 

In this paper, we propose a deep learning–based framework to predict clinically relevant postoperative pancreatic fistula using preoperative CT imaging. We leveraged a prospectively maintained dataset from Mayo Clinic comprising segmented pancreas CT scans and associated surgical outcomes, which, to our knowledge, represents one of the largest imaging cohorts studied for this problem\cite{muaddi2021postoperative}. Our approach uniquely focuses on the pancreas region via auto-segmentation, aiming to enhance the model’s ability to learn relevant morphological and textural features. A suite of state-of-the-art 3D convolutional neural network architectures was implemented and evaluated, including variations of ResNet, and spatiotemporal models (such as R2Plus1D~\cite{tran2018closer} and MC3ResNet18~\cite{tran2018closer}) that capture volumetric contextual information. We also introduce a comprehensive model comparison and evaluation framework: all models were trained under identical conditions and five-fold cross-validation to ensure fair performance assessment. Class imbalance was addressed through balanced sampling techniques, and performance metrics such as the Area Under the Receiver Operating Characteristic Curve (AUC-ROC)  were used to benchmark the algorithms.

\begin{figure}[t!]
    \centering

    \subfloat[]{%
        \includegraphics[
            width=0.76\linewidth,
            trim={0.0cm 0.0cm 0.0cm 0.0cm},
            clip
        ]{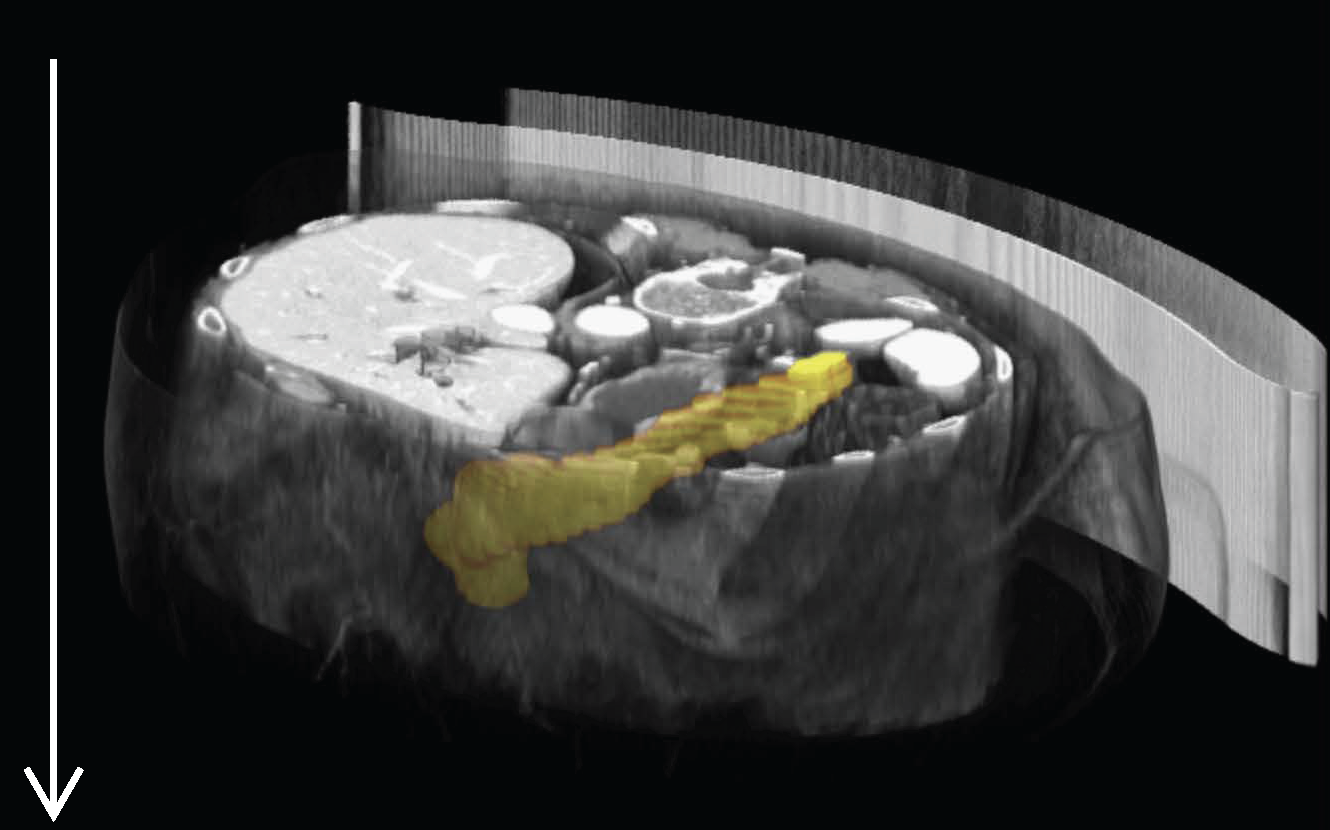}
        \label{fig:raw_ct_slices_left}
    }\hspace{0.1em}%
  
    \subfloat[]{%
        \includegraphics[
            width=0.155\linewidth,
            trim={0.0cm 0cm 0.5cm 4.5cm},
            clip
        ]{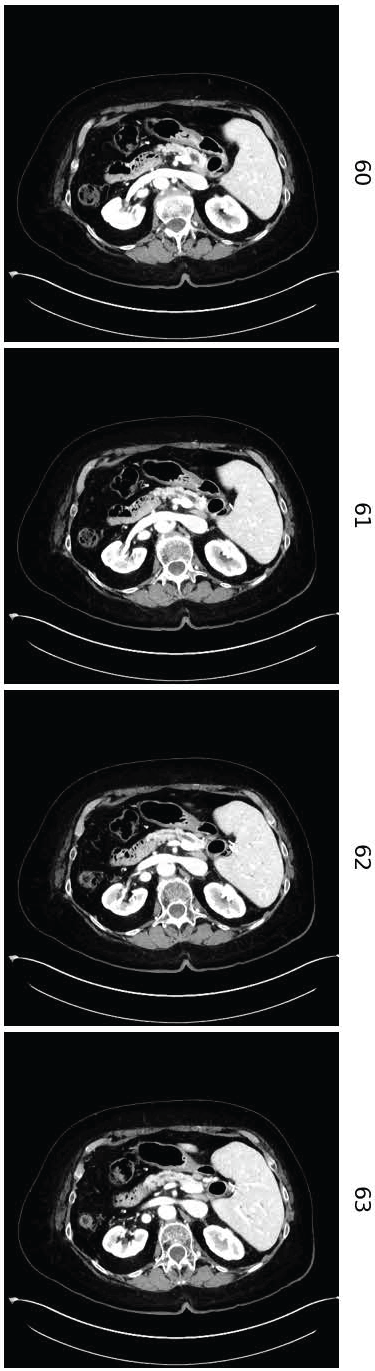}
        \label{fig:raw_ct_slices_right}
    }

    \caption{Raw preoperative CT axial slices illustrating variability in field-of-view and intensity distribution prior to preprocessing.}
    \label{fig:raw_ct_slices}
\end{figure}

\begin{figure}[t!]
    \centering
    \includegraphics[
        width=\linewidth,
        trim={13.0cm 0cm 0cm 1.0cm}, 
        clip
    ]{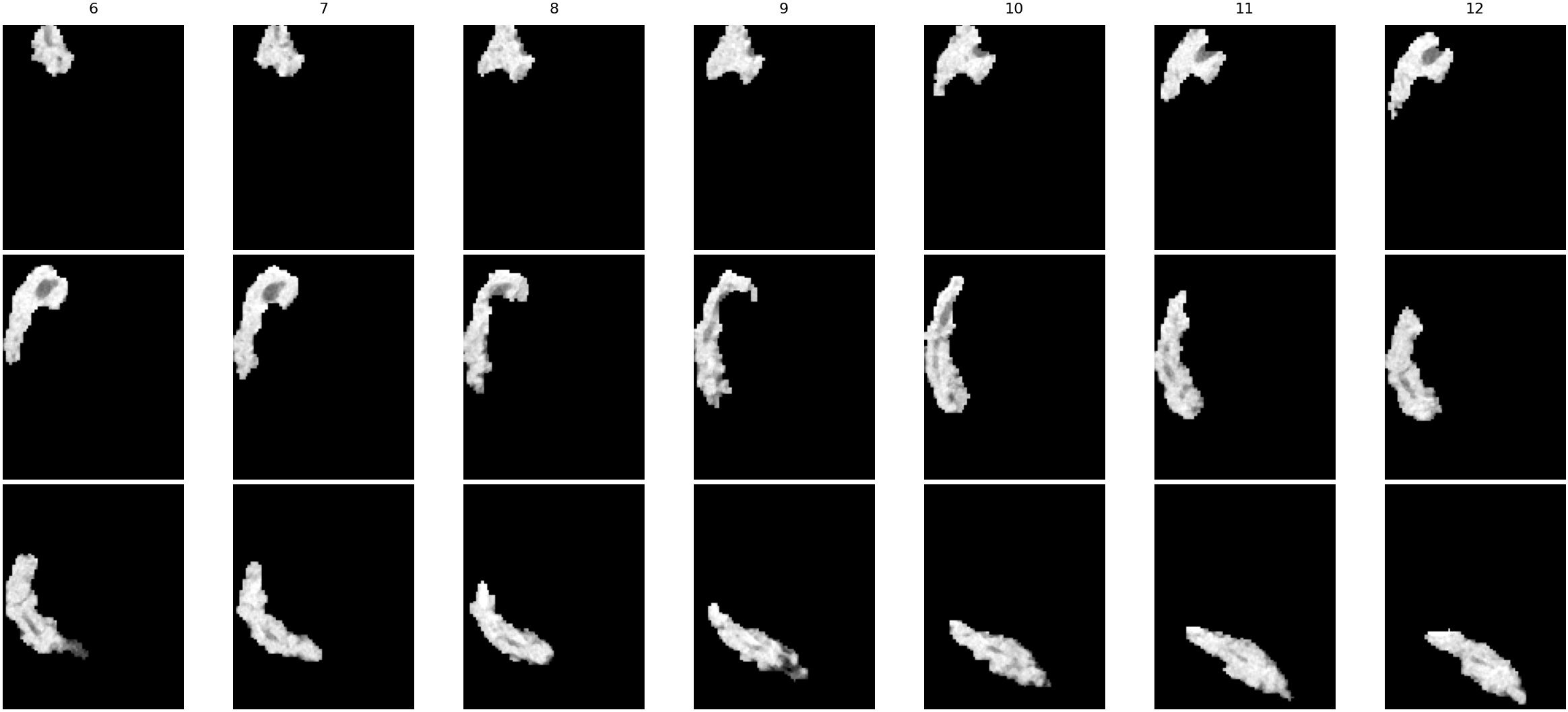}
    \caption{Axial slices after preprocessing. Intensities are windowed in Hounsfield units and volumes are cropped to a pancreas-centric ROI.}
    \label{fig:preproc_ct_slices}
\end{figure}

Our results demonstrate that deep learning models can achieve promising AUC in discriminating patients who will develop POPF from those who will not, based soley on preoperative imaging data. In particular, the top-performing model in our experiments – a 3D ResNet-based architecture with spatiotemporal feature encoding – achieved robust discrimination of high-risk cases, outperforming baseline approaches. These findings highlight the feasibility of preoperative CT-based risk stratification for pancreatic fistula, which could be incorporated into surgical planning workflows. By identifying high-risk patients before surgery, clinicians may alter their clinical decision making, such as avoiding the pancreatic anastamosis altogether which leads to POPF upon breakdown, or more aggressively utilise prophylactic measures, or institute enhanced postoperative monitoring for high-risk individuals. In summary, this work demonstrates the clinical utility of advanced deep learning for perioperative decision support and paves the way for the further development of imaging-driven predictive models in pancreatic surgery. The proposed methodology and evaluation framework can also serve as a foundation for future studies on surgical outcome prediction using multimodal data.

\section{Fistula Classification Model}
An overview of the full processing and classification workflow is shown in Fig~\ref{fig:model_pipeline}. CT scans are separated based on the contrast phase, as determined by a pre-existing XGBoost classifier~\cite{wasserthal2023totalsegmentator}. Commonly used phases for pancreas interrogation on CT scans include late arterial and portal venous phases. Subsequent analyses were therefore stratified based on contrast phase. For each preoperative CT volume, we first apply an automatic pancreas segmentation method to obtain a binary pancreas mask. Using this mask, we isolate the pancreas region via masking and tight bounding-box cropping, followed by center padding to a common volume size. We then apply Hounsfield Unit (HU) windowing to standardize intensities and retain clinically relevant contrast. The resulting preprocessed 3D volumes are finally input to a binary classification network that predicts the presence ($y{=}1$) or absence ($y{=}0$) of postoperative pancreatic fistula (POPF).

\label{sec:mask_pipeline}

\subsection{Preprocessing of CT Images}

Let $\mathcal{D}=\{(\mathbf{V}_i,\mathbf{M}_i,y_i)\}_{i=1}^{N}$ denote a dataset of preoperative CT volumes, corresponding pancreas masks, and binary labels for clinically relevant POPF. Here, $\mathbf{V}_i\in\mathbb{R}^{\mathrm{H}_{\mathrm{in}}\times\mathrm{W}_{\mathrm{in}}\times\mathrm{D}_{\mathrm{in}}}$ is a 3D CT volume, $\mathbf{M}_i\in\{0,1\}^{\mathrm{H}_{\mathrm{in}}\times\mathrm{W}_{\mathrm{in}}\times\mathrm{D}_{\mathrm{in}}}$ is its associated binary mask, and $y_i\in\{0,1\}$ is the outcome label. We define the support of the pancreas mask as
\begin{equation}
\operatorname{supp}(\mathbf{M}_i)
=\{(r,s,t)\mid \mathbf{M}_i(r,s,t)=1\}.
\end{equation}
We compute the tight axis-aligned bounding box
\begin{equation}
\begin{aligned}
B_i&=\bigl[r_i^{\min}\!:\!r_i^{\max}\bigr]\times
     \bigl[s_i^{\min}\!:\!s_i^{\max}\bigr]\times
     \bigl[t_i^{\min}\!:\!t_i^{\max}\bigr],
\end{aligned}
\end{equation}
which encloses $\operatorname{supp}(\mathbf{M}_i)$. The masked and cropped volume is
\begin{equation}
\mathbf{U}_i=\operatorname{crop}\bigl(\mathbf{V}_i\odot \mathbf{M}_i,\; B_i\bigr)\in\mathbb{R}^{b_i^r\times b_i^s\times b_i^t},
\end{equation}
where $\odot$ denotes element-wise multiplication, $(b_i^r,b_i^s,b_i^t)$ are the side lengths of $B_i$, and $\operatorname{crop}(\mathbf{A},B)$ returns the subvolume of $\mathbf{A}$ indexed by the axis-aligned box $B$.
To obtain a common input size, we define the global maximum box as
\begin{equation}
\bigl(\mathrm{H}_{\mathrm{in}}^{\star},\,\mathrm{W}_{\mathrm{in}}^{\star},\,\mathrm{D}_{\mathrm{in}}^{\star}\bigr)
=\max_{i}\,\bigl(b_i^r,\,b_i^s,\,b_i^t\bigr)
\qquad \text{(componentwise)}.
\end{equation}
Each cropped volume $\mathbf{U}_i$ is then center-padded to this size using its minimum voxel value, yielding
\begin{equation}
\tilde{\mathbf{U}}_i=\operatorname{pad}_{\bigl(\mathrm{H}_{\mathrm{in}}^{\star},\,\mathrm{W}_{\mathrm{in}}^{\star},\,\mathrm{D}_{\mathrm{in}}^{\star}\bigr)}^{\mathrm{center}}(\mathbf{U}_i)\in\mathbb{R}^{\mathrm{H}_{\mathrm{in}}^{\star}\times\mathrm{W}_{\mathrm{in}}^{\star}\times\mathrm{D}_{\mathrm{in}}^{\star}}.
\end{equation}
We then apply the HU windowing operator $\mathcal{W}_{w,c}(\cdot)$, parameterized by window width $w$ and center $c$ (default $w{=}300$, $c{=}80$), which maps intensities to $[0,255]$ according to
\begin{equation}
\mathcal{W}_{w,c}(v)=255\cdot\frac{\min\{\max\{v,\,c-\tfrac{w}{2}\},\,c+\tfrac{w}{2}\}-\left(c-\tfrac{w}{2}\right)}{w}.
\end{equation}
The resulting network input is the windowed volume
\begin{equation}
\mathbf{X}_i=\mathcal{W}_{w,c}(\tilde{\mathbf{U}}_i)\in\mathbb{R}^{\mathrm{H}_{\mathrm{in}}^{\star}\times\mathrm{W}_{\mathrm{in}}^{\star}\times\mathrm{D}_{\mathrm{in}}^{\star}}.
\end{equation}

\subsection{Classification Models}
\label{sec:classification_models}
After preprocessing, each CT case is represented by a fixed-size 3D input volume
$\mathbf{X}_i \in \mathbb{R}^{\mathrm{H}_{\mathrm{in}}^{\star}\times\mathrm{W}_{\mathrm{in}}^{\star}\times\mathrm{D}_{\mathrm{in}}^{\star}}$.
We evaluate three 3D classification backbones, denoted by
$\{g_{\theta_m}^{(m)}\}_{m=1}^{3}$, for binary POPF prediction, where
$g_{\theta_m}^{(m)}$ denotes the $m$th model with parameters $\theta_m$.
Given $\mathbf{X}_i$, the $m$th model produces a logit vector
\begin{equation}
\mathbf{z}_i^{(m)} = g_{\theta_m}^{(m)}(\mathbf{X}_i)\in\mathbb{R}^{2},
\end{equation}
which is converted into posterior class probabilities via
\begin{equation}
\hat{\mathbf{p}}_i^{(m)}=\mathrm{softmax}\!\left(\mathbf{z}_i^{(m)}\right).
\end{equation}

\subsubsection{CNN3D}
\label{sec:cnn3d}
CNN3D is a compact 3D convolutional network trained from scratch. It consists of three convolutional stages with channel widths $16\!\rightarrow\!32\!\rightarrow\!64$, each using $3\times3\times3$ kernels, BatchNorm, and ReLU activations, with $2\times2\times2$ max-pooling applied after each stage. The second stage incorporates a residual shortcut implemented by a $1\times1\times1$ convolution to match the change in feature dimension from $16$ to $32$. Let $\sigma(\cdot)$ denote the ReLU activation and let $\mathrm{MP}(\cdot)$ denote $2\times2\times2$ max-pooling. The feature extraction pipeline can be written as
\begin{align}
\mathbf{u}_i^{(1)} &= \mathrm{MP}\!\left(\sigma\!\left(\mathrm{BN}_1(\mathrm{Conv}_1(\mathbf{X}_i))\right)\right), \\
\tilde{\mathbf{u}}_i^{(2)} &= \sigma\!\left(\mathrm{BN}_{2b}\!\left(\mathrm{Conv}_{2b}\!\left(\sigma\!\left(\mathrm{BN}_{2a}(\mathrm{Conv}_{2a}(\mathbf{u}_i^{(1)}))\right)\right)\right)\right), \\
\mathbf{u}_i^{(2)} &= \mathrm{MP}\!\left(\tilde{\mathbf{u}}_i^{(2)} + \mathrm{Conv}_{\text{short}}(\mathbf{u}_i^{(1)})\right), \\
\mathbf{u}_i^{(3)} &= \mathrm{MP}\!\left(\sigma\!\left(\mathrm{BN}_3(\mathrm{Conv}_3(\mathbf{u}_i^{(2)}))\right)\right),
\end{align}
where $\mathrm{Conv}_{\text{short}}$ denotes the $1\times1\times1$ shortcut convolution. Applying adaptive global average pooling, denoted by $\mathrm{GAP}(\cdot)$, reduces the spatial dimensions to $1\times1\times1$ and yields the embedding
\begin{equation}
\mathbf{h}_i = \mathrm{GAP}\!\left(\mathbf{u}_i^{(3)}\right)\in\mathbb{R}^{64}.
\end{equation}
Finally, a two-layer fully connected classification head with dropout ($p{=}0.2$) produces the logits
\begin{equation}
\mathbf{z}_i^{(\mathrm{cnn})}
= \mathbf{W}_2\,\mathrm{Drop}\!\left(\sigma(\mathbf{W}_1\mathbf{h}_i+\mathbf{b}_1)\right) + \mathbf{b}_2 \in \mathbb{R}^{2}.
\end{equation}

\subsubsection{ResNet-(2+1)D-18}
\label{sec:r2plus1d}
ResNet-(2+1)D-18 is a residual 3D architecture that factorizes a 3D convolution into a spatial convolution followed by a depth (inter-slice) convolution, enabling efficient modeling of 3D context~\cite{tran2018closer}. The backbone outputs a pooled feature vector $\mathbf{h}_i\in\mathbb{R}^{D}$, which is mapped to logits by a linear layer as

\begin{equation}
\mathbf{z}_i^{(\mathrm{r2{+}1d})}
= \mathbf{W}^{(\mathrm{r2{+}1d})}\mathbf{h}_i + \mathbf{b}^{(\mathrm{r2{+}1d})}.
\end{equation} 

When initializing from pretrained video weights (three-channel input), we replicate the single-channel windowed CT input before forwarding.

\subsubsection{ResNet-MC3-18}
\label{sec:mc3}
ResNet-MC3-18 ~\cite{tran2018closer} uses \emph{mixed} spatiotemporal convolutions: early layers use 3D kernels to capture inter-slice context, while later layers use predominantly slice-wise spatial convolutions (depth kernel size $1$) to reduce complexity. The backbone produces a pooled embedding $\mathbf{h}_i\in\mathbb{R}^{D}$ followed by a linear classifier defined as
\begin{equation}
\mathbf{z}_i^{(\mathrm{mc3})}
= \mathbf{W}^{(\mathrm{mc3})}\mathbf{h}_i + \mathbf{b}^{(\mathrm{mc3})}.
\end{equation}
For pretrained initialization, we again replicate the single-channel windowed CT input to match the expected three-channel interface.

\begin{table}[t!]
\centering
\caption{Cohort characteristics by CR-POPF.}
\label{tab:demographics_combined}
\scriptsize
\setlength{\tabcolsep}{2pt}
\renewcommand{\arraystretch}{1.00}

\begin{tabularx}{\columnwidth}{@{}>{\raggedright\arraybackslash}p{0.56\columnwidth} r r r@{}}
\toprule
 & \textbf{No} & \textbf{Yes} & \textbf{All}  \\
\midrule

\multicolumn{4}{@{}l}{\textbf{Portal venous cohort (N=1353)}} \\
N & 1160 & 193 & 1353  \\
BMI (kg/m$^2$) & 26.6$\pm$5.2 & 28.5$\pm$4.8 & 26.8$\pm$5.2  \\
\addlinespace[0.5ex]
\multicolumn{4}{@{}l}{\textit{Tumor type, N (\%) }} \\
\quad PDAC & 621 (53.5) & 59 (30.6) & 680 (50.3) \\
\quad Neuroendocrine tumor (NET) & 81 (7.0) & 43 (22.3) & 124 (9.2) \\
\quad IPMN (non-invasive) & 93 (8.0) & 21 (10.9) & 114 (8.4) \\
\quad Other malignant & 249 (21.5) & 49 (25.4) & 298 (22.0) \\
\quad Other benign & 59 (5.1) & 17 (8.8) & 76 (5.6) \\
\quad Pancreatitis/infect./inflam. & 50 (4.3) & 3 (1.6) & 53 (3.9) \\
\quad Other & 7 (0.6) & 1 (0.5) & 8 (0.6) \\

\midrule

\multicolumn{4}{@{}l}{\textbf{Late arterial cohort (N=963)}} \\
N & 834 & 129 & 963  \\
BMI (kg/m$^2$) & 26.8$\pm$5.2 & 28.8$\pm$4.8 & 27.1$\pm$5.2  \\
\addlinespace[0.5ex]
\multicolumn{4}{@{}l}{\textit{Tumor type, N (\%) }} \\
\quad PDAC & 502 (60.2) & 46 (35.7) & 548 (56.9) \\
\quad Neuroendocrine tumor (NET) & 58 (7.0) & 33 (25.6) & 91 (9.4) \\
\quad IPMN (non-invasive) & 65 (7.8) & 12 (9.3) & 77 (8.0) \\
\quad Other malignant & 140 (16.8) & 25 (19.4) & 165 (17.1) \\
\quad Other benign & 36 (4.3) & 11 (8.5) & 47 (4.9) \\
\quad Pancreatitis/infect./inflam. & 30 (3.6) & 2 (1.6) & 32 (3.3) \\
\quad Other & 3 (0.4) & 0 (0.0) & 3 (0.3) \\

\bottomrule
\end{tabularx}
\end{table}

\subsection{Data Augmentation}
During training, each input $\mathbf{X}_i$ (windowed HU volume) is perturbed on-the-fly by a stochastic MONAI transform pipeline~\cite{cardoso2022monai} to improve robustness to anatomical pose, minor misregistration, and acquisition variability. The pipeline samples from spatial transforms (random 3D flips, small-angle rotations, and mild zoom) and intensity transforms (histogram/contrast adjustments), and additionally applies low-amplitude Gaussian noise or random smoothing/sharpening. Each operator is applied independently with predefined probabilities, yielding diverse yet anatomically plausible variants of the \emph{windowed} ROI volumes.

\subsection{Optimization and Scheduling}
We fine-tune the ResNet-(2+1)D-18 and ResNet-MC3-18 by separating parameters into two groups: the pretrained backbone $\theta_{\text{back}}$ and the classification head $\theta_{\text{head}}$, and assign group-specific learning rates $(\eta_{\text{back}},\eta_{\text{head}})$, typically with $\eta_{\text{head}}>\eta_{\text{back}}$ to adapt the new head more rapidly. Optimization is performed with AdamW~\cite{loshchilov2017decoupled}.

\begin{figure*}[t!]
    \centering

    \makebox[\textwidth][c]{%
        \subfloat[\textbf{MS}—Late arterial]{%
            \includegraphics[width=0.235\textwidth]{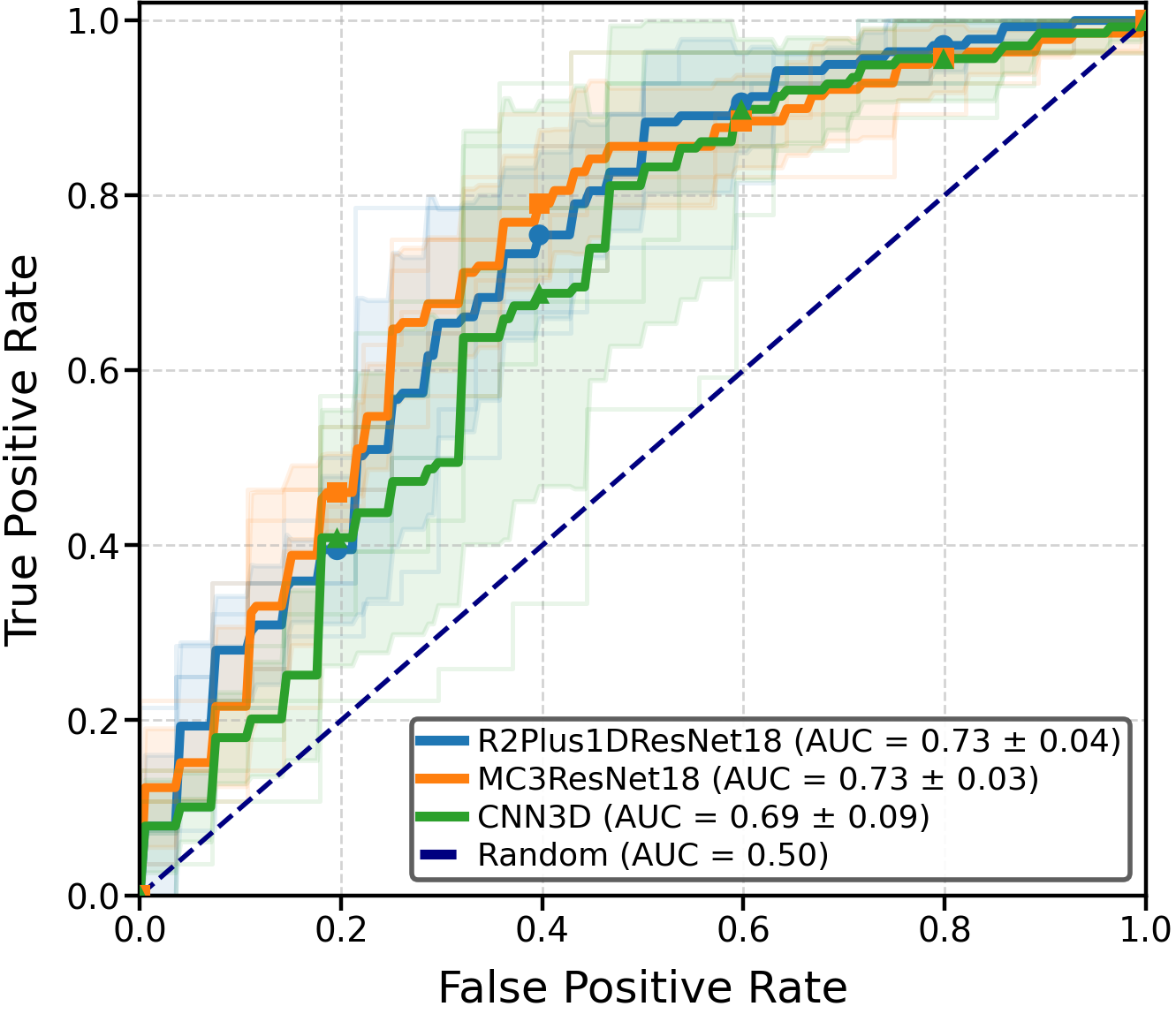}%
        }%
        \hspace{0.01\textwidth}%
        \subfloat[\textbf{MS}—Portal venous]{%
            \includegraphics[width=0.235\textwidth]{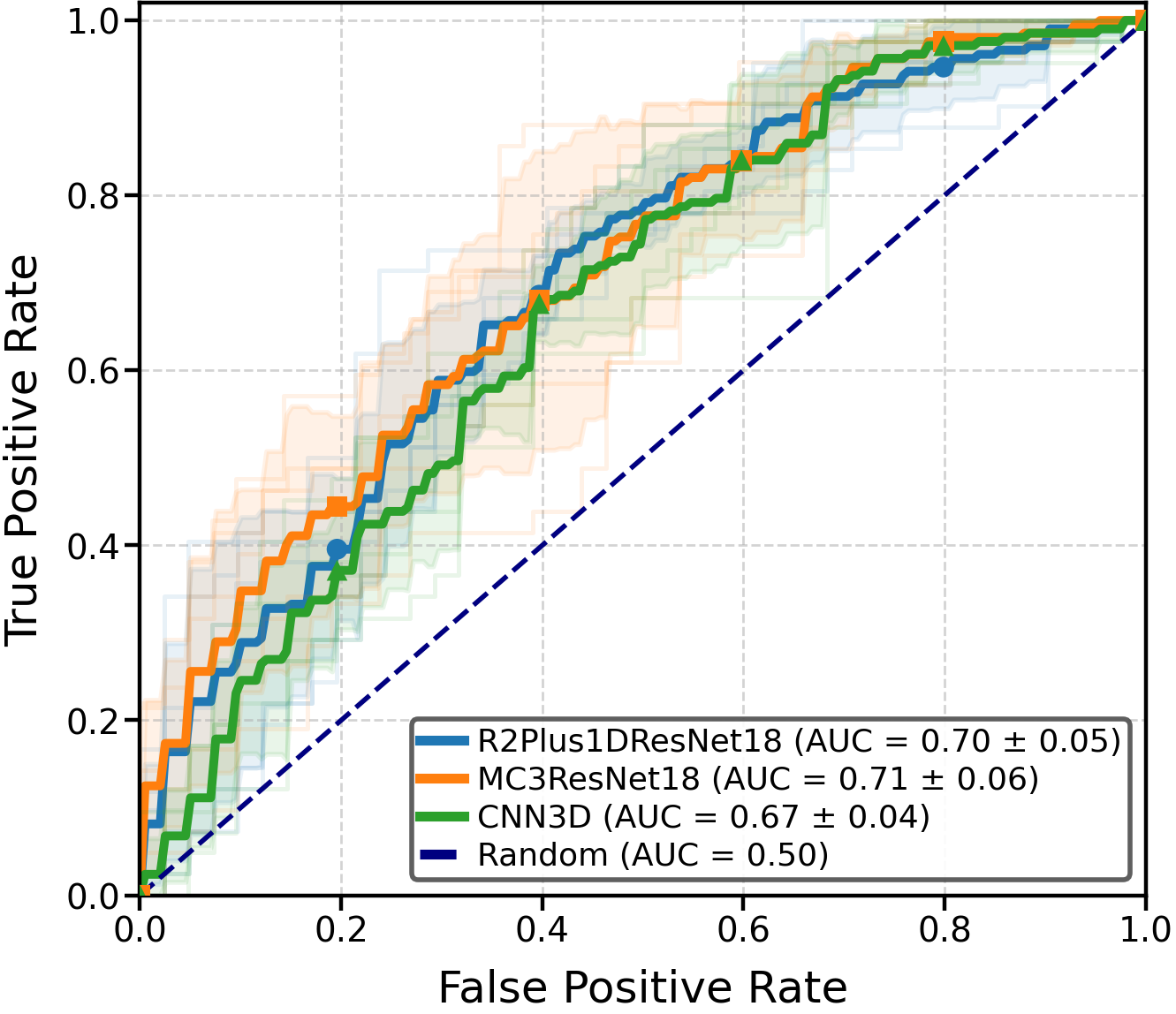}%
        }%
        \hspace{0.01\textwidth}%
        \subfloat[\textbf{TS}—Late arterial]{%
            \includegraphics[width=0.235\textwidth]{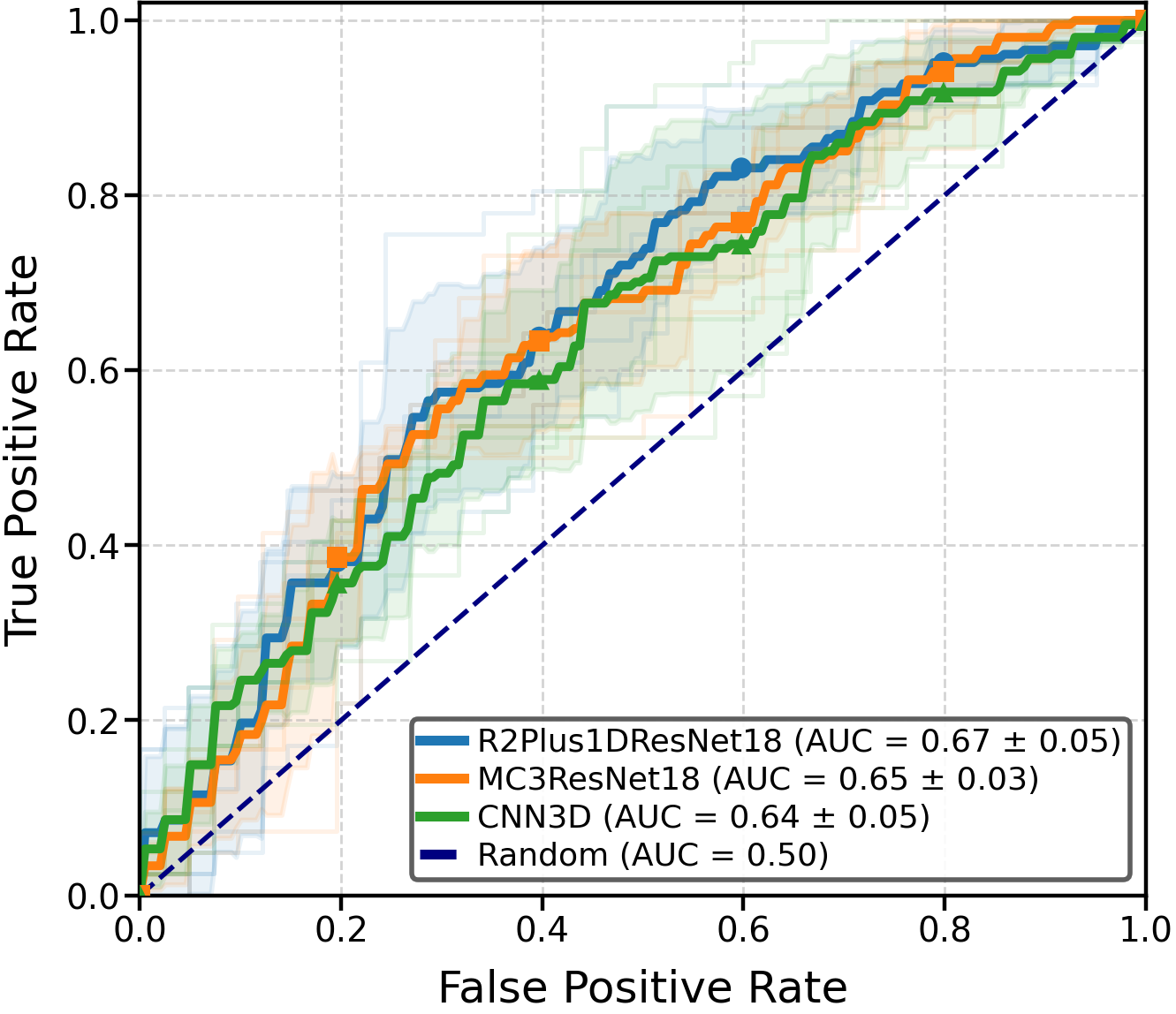}%
        }%
        \hspace{0.01\textwidth}%
        \subfloat[\textbf{TS}—Portal venous]{%
            \includegraphics[width=0.235\textwidth]{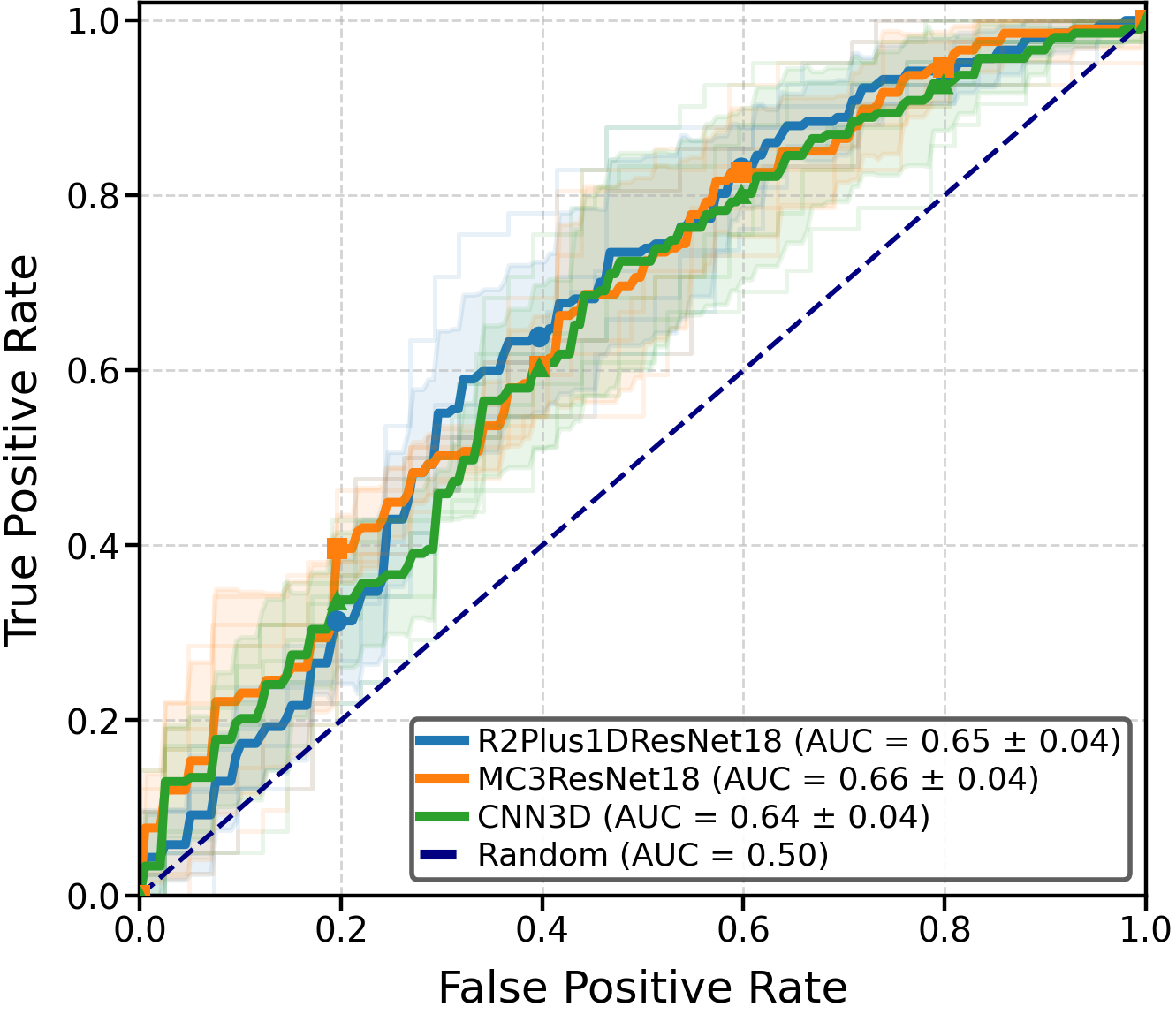}%
        }%
    }

    \caption{ROC curves for POPF prediction on balanced splits comparing MS vs.\ TS pancreas masks across late arterial and portal venous cohorts.}
    \label{fig:auc_scores_pretrained}
\end{figure*}

\begin{table*}[t!]
    \centering
    \caption{Classification performance results of the fine-tuned 3D models using balanced train/test sets. MS: Mayo Segmentation, TS: TotalSegmentator, Seg: Segmentation type.}
    \label{tab:performance_comparison_3d_best_bal_dist}
    \scriptsize
    \setlength{\tabcolsep}{3pt}
    \renewcommand{\arraystretch}{1.12}

    \begin{adjustbox}{max width=\textwidth}
    \begin{tabular}{@{}llccccc ccccc@{}}
        \toprule
        \multirow{2}{*}{Model} &
        \multirow{2}{*}{Seg.} &
        \multicolumn{5}{c}{Arterial} &
        \multicolumn{5}{c}{Portal Venous} \\
        \cmidrule(lr){3-7}\cmidrule(lr){8-12}
        & & AUC & Acc. & Prec. & Rec. & F1 & AUC & Acc. & Prec. & Rec. & F1 \\
        \midrule

        \multirow{2}{*}{CNN3D}
            & MS & $0.68\!\pm\!0.08$ & $0.70\!\pm\!0.06$ & $0.64\!\pm\!0.05$ & $0.89\!\pm\!0.04$ & $0.75\!\pm\!0.05$
                 & $0.67\!\pm\!0.03$ & $0.66\!\pm\!0.02$ & $0.64\!\pm\!0.03$ & $0.75\!\pm\!0.16$ & $0.68\!\pm\!0.06$ \\
            & TS & $0.63\!\pm\!0.05$ & $0.64\!\pm\!0.04$ & $0.67\!\pm\!0.08$ & $0.62\!\pm\!0.23$ & $0.62\!\pm\!0.12$
                 & $0.64\!\pm\!0.03$ & $0.64\!\pm\!0.04$ & $0.63\!\pm\!0.05$ & $0.72\!\pm\!0.21$ & $0.65\!\pm\!0.10$ \\
        \addlinespace[2pt]

        \multirow{2}{*}{R2Plus1DResNet18}
            & MS & $\boldsymbol{0.73\!\pm\!0.04}$ & $0.72\!\pm\!0.04$ & $0.70\!\pm\!0.06$ & $0.81\!\pm\!0.10$ & $0.75\!\pm\!0.05$
                 & $\boldsymbol{0.73\!\pm\!0.06}$ & $0.72\!\pm\!0.04$ & $0.71\!\pm\!0.07$ & $0.76\!\pm\!0.09$ & $0.73\!\pm\!0.03$ \\
            & TS & $0.66\!\pm\!0.04$ & $0.67\!\pm\!0.04$ & $0.66\!\pm\!0.05$ & $0.70\!\pm\!0.13$ & $0.67\!\pm\!0.06$
                 & $0.65\!\pm\!0.03$ & $0.65\!\pm\!0.04$ & $0.65\!\pm\!0.02$ & $0.65\!\pm\!0.12$ & $0.65\!\pm\!0.06$ \\
        \addlinespace[2pt]

        \multirow{2}{*}{MC3ResNet18}
            & MS & $\boldsymbol{0.73\!\pm\!0.02}$ & $0.72\!\pm\!0.03$ & $0.74\!\pm\!0.04$ & $0.71\!\pm\!0.17$ & $0.72\!\pm\!0.08$
                 & $0.71\!\pm\!0.06$ & $0.69\!\pm\!0.04$ & $0.72\!\pm\!0.08$ & $0.69\!\pm\!0.21$ & $0.68\!\pm\!0.07$ \\
            & TS & $0.65\!\pm\!0.03$ & $0.65\!\pm\!0.03$ & $0.68\!\pm\!0.04$ & $0.60\!\pm\!0.13$ & $0.63\!\pm\!0.06$
                 & $0.65\!\pm\!0.03$ & $0.65\!\pm\!0.03$ & $0.68\!\pm\!0.10$ & $0.65\!\pm\!0.24$ & $0.63\!\pm\!0.12$ \\
        \bottomrule
    \end{tabular}
    \end{adjustbox}
\end{table*}

\section{EXPERIMENTS}

\subsection{Dataset}
Patients were identified from a prospectively maintained registry of Mayo Clinic pancreatic operations. We included all adult patients (age $\geq$ 18 years) who underwent pancreatoduodenectomy for any indication between 2010 and 2020, had a preoperative computed tomography (CT) scan available, and had a known postoperative pancreatic fistula (POPF) outcome. For each patient, the CT study closest to the date of surgery was retrieved. Patients who did not provide research authorization were excluded, as were patients with missing POPF outcome data.

Two CT phase-specific cohorts were analyzed: (i) a portal venous cohort ($N{=}1353$) and (ii) a late arterial cohort ($N{=}963$). Baseline demographics and clinical characteristics stratified by clinically relevant POPF (CR-POPF)\cite{bassi20172016} are summarized in Table~\ref{tab:demographics_combined}.

\begin{figure*}[t!]
    \centering

    \makebox[\textwidth][c]{%
        \subfloat[\textbf{MS}—Late arterial]{%
            \includegraphics[width=0.235\textwidth]{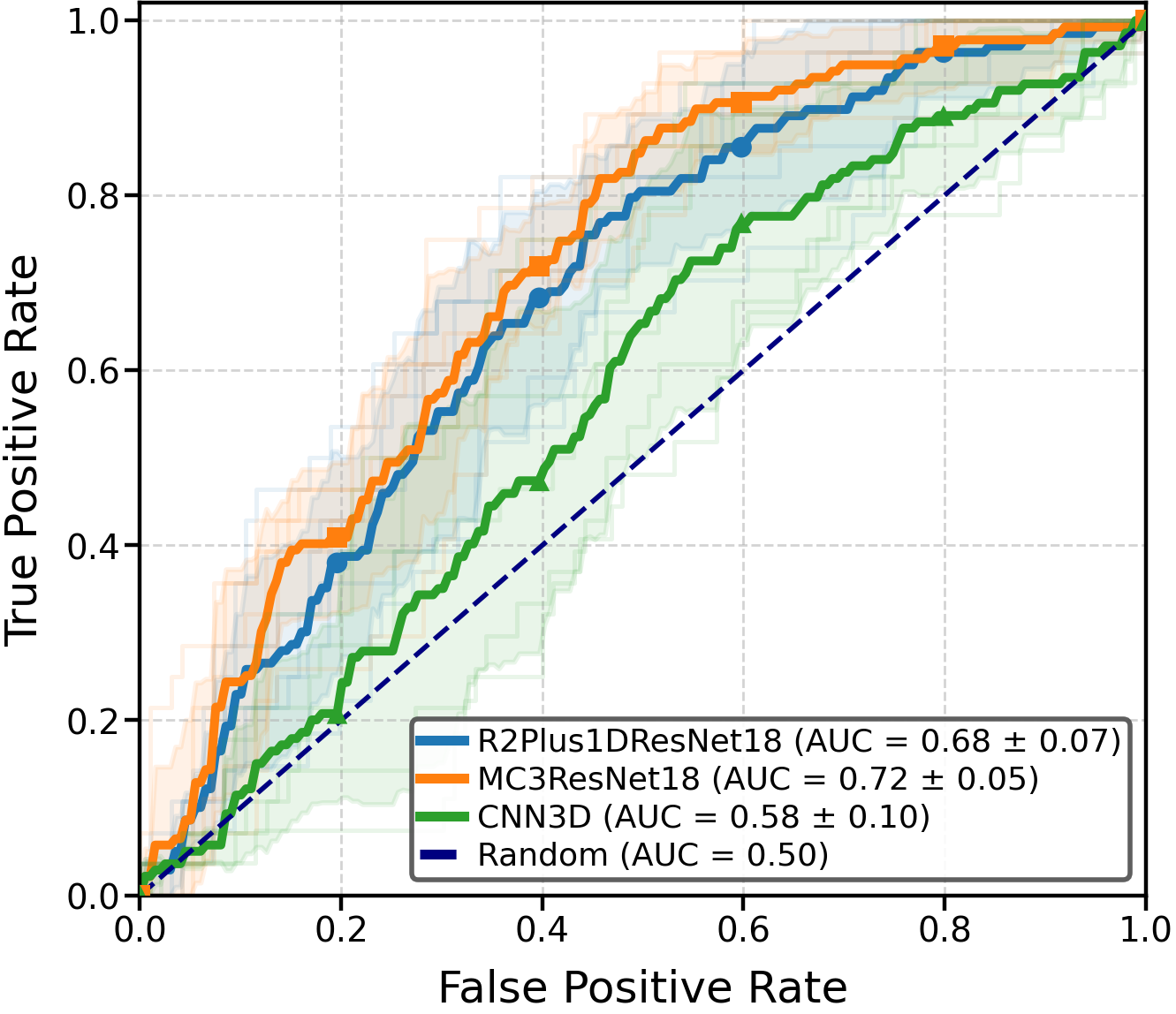}%
        }%
        \hspace{0.01\textwidth}%
        \subfloat[\textbf{MS}—Portal venous]{%
            \includegraphics[width=0.235\textwidth]{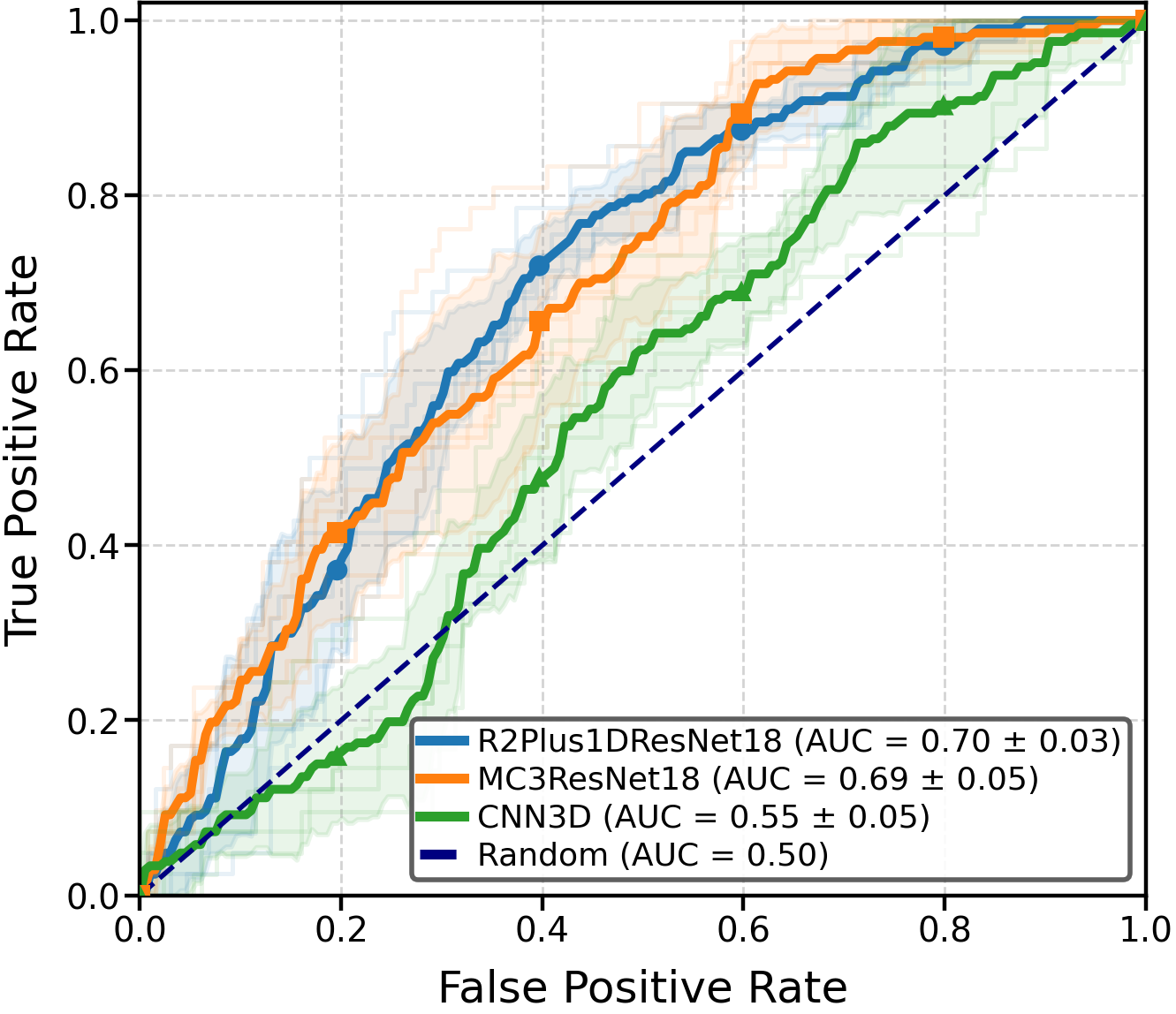}%
        }%
        \hspace{0.01\textwidth}%
        \subfloat[\textbf{TS}—Late arterial]{%
            \includegraphics[width=0.235\textwidth]{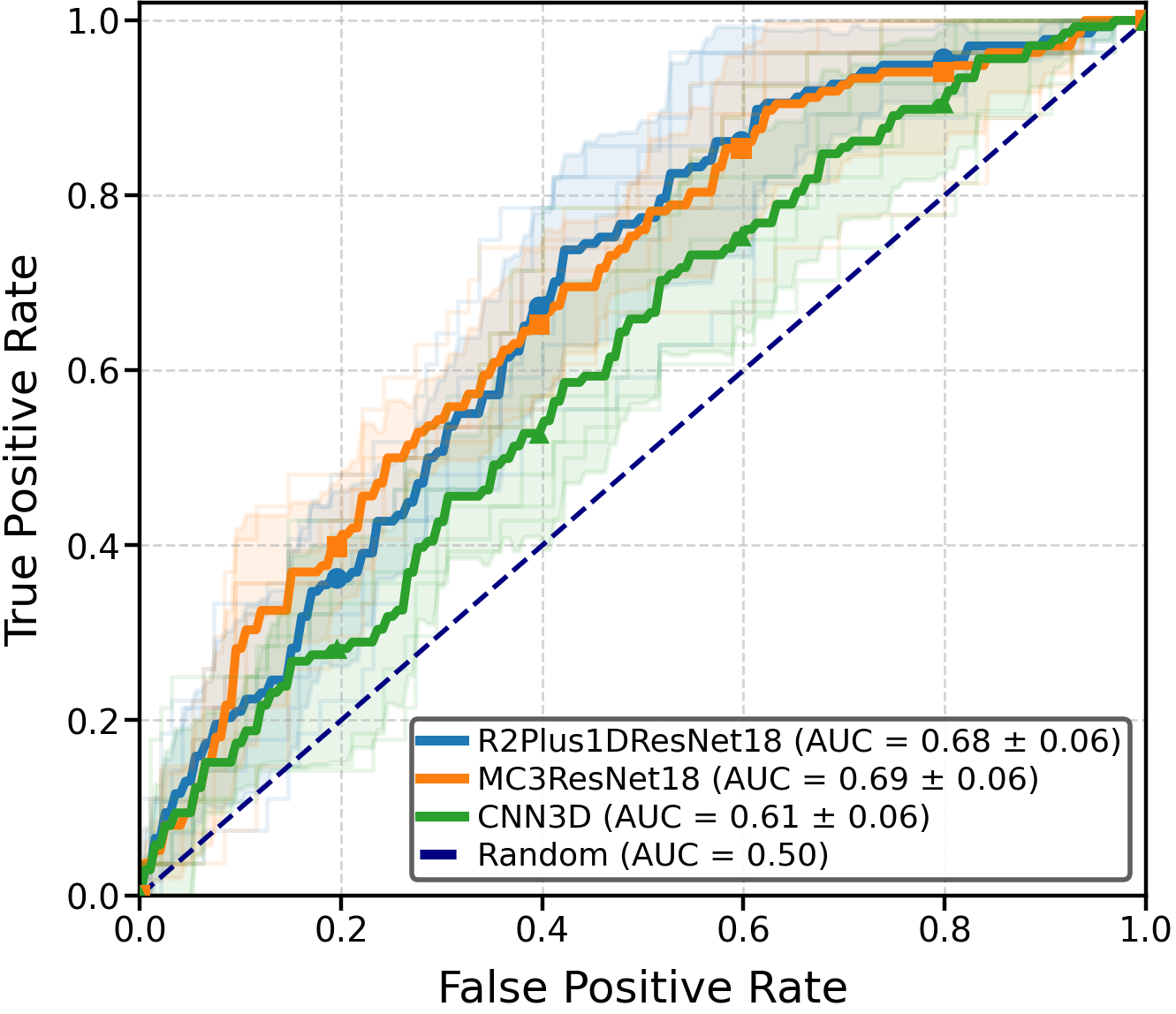}%
        }%
        \hspace{0.01\textwidth}%
        \subfloat[\textbf{TS}—Portal venous]{%
            \includegraphics[width=0.235\textwidth]{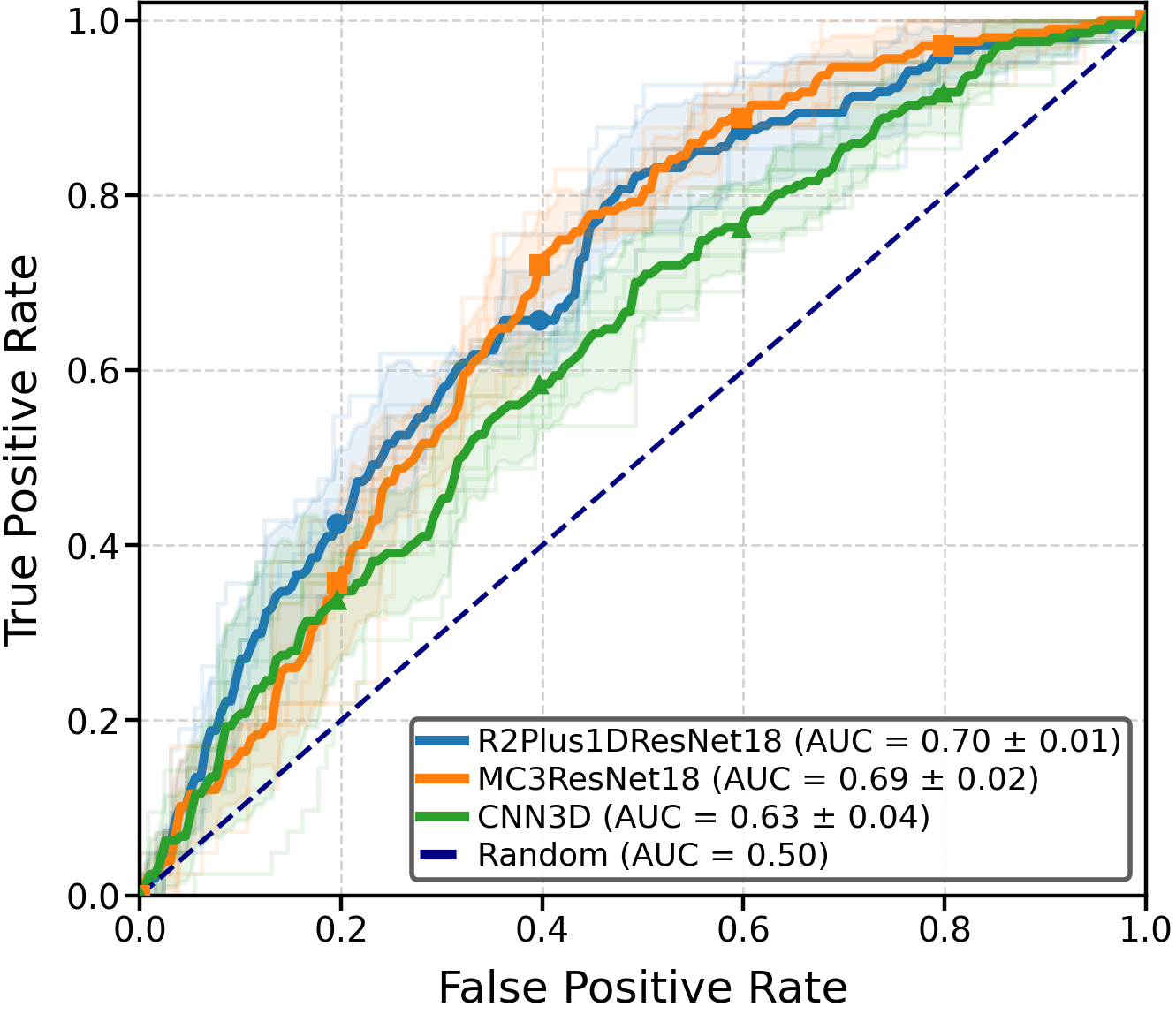}%
        }%
    }

    \caption{ROC curves for POPF prediction using natural-prevalence splits with MS vs.\ TS pancreas ROIs across late arterial and portal venous cohorts.}

    \label{fig:auc_scores_trained_natural_dist}
\end{figure*}

\begin{table*}[t!]
    \centering
    \caption{Classification performance results of the fine-tuned 3D models using natural distribution in train/test sets. MS: Mayo Segmentation, TS: TotalSegmentator, Seg: Segmentation type.}
    \label{tab:performance_comparison_3d_best_natural_dist}
    \scriptsize
    \setlength{\tabcolsep}{3pt}
    \renewcommand{\arraystretch}{1.12}

    \begin{adjustbox}{max width=\textwidth}
    \begin{tabular}{@{}llccccc ccccc@{}}
        \toprule
        \multirow{2}{*}{Model} &
        \multirow{2}{*}{Seg.} &
        \multicolumn{5}{c}{Arterial} &
        \multicolumn{5}{c}{Portal Venous} \\
        \cmidrule(lr){3-7}\cmidrule(lr){8-12}
        & & AUC & Acc. & Prec. & Rec. & F1 & AUC & Acc. & Prec. & Rec. & F1 \\
        \midrule

        \multirow{2}{*}{CNN3D}
            & MS & $0.58\!\pm\!0.10$ & $0.49\!\pm\!0.08$ & $0.28\!\pm\!0.04$ & $0.81\!\pm\!0.12$ & $0.42\!\pm\!0.06$
                 & $0.56\!\pm\!0.05$ & $0.49\!\pm\!0.07$ & $0.29\!\pm\!0.02$ & $0.78\!\pm\!0.15$ & $0.42\!\pm\!0.03$ \\
            & TS & $0.61\!\pm\!0.06$ & $0.56\!\pm\!0.09$ & $0.31\!\pm\!0.03$ & $0.76\!\pm\!0.20$ & $0.43\!\pm\!0.05$
                 & $0.63\!\pm\!0.04$ & $0.61\!\pm\!0.10$ & $0.37\!\pm\!0.09$ & $0.65\!\pm\!0.16$ & $0.45\!\pm\!0.02$ \\
        \addlinespace[2pt]

        \multirow{2}{*}{R2Plus1DResNet18}
            & MS & $0.68\!\pm\!0.07$ & $0.61\!\pm\!0.06$ & $0.35\!\pm\!0.04$ & $0.82\!\pm\!0.12$ & $0.49\!\pm\!0.05$
                 & $\boldsymbol{0.70\!\pm\!0.03}$ & $0.64\!\pm\!0.05$ & $0.38\!\pm\!0.05$ & $0.75\!\pm\!0.05$ & $0.50\!\pm\!0.03$ \\
            & TS & $0.68\!\pm\!0.06$ & $0.58\!\pm\!0.07$ & $0.34\!\pm\!0.04$ & $0.86\!\pm\!0.10$ & $0.48\!\pm\!0.05$
                 & $\boldsymbol{0.70\!\pm\!0.01}$ & $0.65\!\pm\!0.07$ & $0.39\!\pm\!0.05$ & $0.76\!\pm\!0.16$ & $0.51\!\pm\!0.02$ \\
        \addlinespace[2pt]

        \multirow{2}{*}{MC3ResNet18}
            & MS & $\boldsymbol{0.72\!\pm\!0.05}$ & $0.62\!\pm\!0.07$ & $0.36\!\pm\!0.05$ & $0.85\!\pm\!0.06$ & $0.50\!\pm\!0.04$
                 & $0.69\!\pm\!0.05$ & $0.62\!\pm\!0.10$ & $0.38\!\pm\!0.06$ & $0.80\!\pm\!0.19$ & $0.50\!\pm\!0.04$ \\
            & TS & $0.69\!\pm\!0.06$ & $0.60\!\pm\!0.08$ & $0.34\!\pm\!0.03$ & $0.80\!\pm\!0.18$ & $0.47\!\pm\!0.04$
                 & $0.69\!\pm\!0.02$ & $0.62\!\pm\!0.05$ & $0.37\!\pm\!0.02$ & $0.82\!\pm\!0.10$ & $0.51\!\pm\!0.01$ \\
        \bottomrule
    \end{tabular}
    \end{adjustbox}
\end{table*}

 The primary outcome was clinically relevant POPF (CR-POPF), defined as International Study Group for Pancreatic Surgery (ISGPS) grade B or C POPF~\cite{bassi2005postoperative}.

\subsection{Phase detection}
To objectively stratify CT scans by contrast timing, we used a TotalSegmentator~\cite{wasserthal2023totalsegmentator}-based phase prediction pipeline in which the median Hounsfield Unit (HU) intensity within each structure is extracted as a feature vector. These features are input to a pretrained XGBoost model that predicts post-injection time and maps it to one of four contrast phases (native, arterial\_early, arterial\_late, portal\_venous).

\subsection{Data preprocessing}
 As illustrated in Fig.~\ref{fig:raw_ct_slices}, raw axial slices exhibit substantial variability in field-of-view, surrounding anatomy, and intensity distribution, which can introduce spurious correlations if used directly for learning.

To focus the model on anatomically relevant information, we defined a patient-specific pancreas region of interest (ROI) using automatically generated pancreas masks. We performed two preprocessing experiments that differed only in the segmentation source used to derive the pancreas mask: (i) a TotalSegmentator-derived pancreas mask (TS)~\cite{wasserthal2023totalsegmentator}, and (ii) a Mayo Segmentation-derived pancreas mask (MS) based on a 3D nnU-Net\cite{isensee2021nnu} pancreas segmentation model trained and externally validated on large CT cohorts \cite{mukherjee2025pancreas}.

For each experiment, each volume was first windowed in Hounsfield units using a soft-tissue window (with intensities clipped and linearly normalized), and then cropped to the tight bounding box of the pancreas. Cropped volumes were padded or center-cropped as needed to obtain a uniform input size for the network. Fig.~\ref{fig:preproc_ct_slices} qualitatively summarizes the effect of windowing and pancreas-centric cropping, which reduces irrelevant background while improving contrast in pancreatic tissue. The same preprocessing pipeline was applied to both late arterial and portal venous phase scans identified in the phase detection step.

\subsection{Training and Evaluation Setup}
We trained \emph{volume-based} 3D classification models for binary POPF prediction from preoperative CT volumes. Samples with missing labels were excluded, and a single-scan-per-patient cohort was constructed. 

Performance was estimated using 5-fold stratified cross-validation with fixed random seeds; per-fold train/validation splits and per-sample predictions were saved. Class imbalance was addressed via weighted sampling during training and a weighted cross-entropy loss. Training used MONAI 3D augmentations (random flips, rotations, zoom, and intensity/noise perturbations), AdamW optimization with a OneCycle learning-rate schedule, and early stopping based on validation ROC-AUC. We report accuracy, precision, recall, F1, balanced accuracy, ROC-AUC averaged across folds.

\subsection{Hardware and Software Environment}
\label{sec:hardware}

Experiments were run on a GPU node with three NVIDIA H100 NVL GPUs (95,830 MiB $\approx$ 94 GB memory per GPU), NVIDIA driver 560.35.05 with CUDA 12.6, and 1.0 TiB system RAM (8.0 GiB swap). The node provided 128 logical CPU cores (2 sockets, 32 cores per socket, 2 threads per core). The default system Python on the node was Python 3.11.7; deep learning libraries (e.g., PyTorch\cite{paszke2019pytorch}/MONAI) were used from the project training environment.

\subsection{Performance Results Analysis}
\label{sec:perf_analysis}

We report cross-validated performance under two evaluation protocols: (i) balanced train/test splits (Table~\ref{tab:performance_comparison_3d_best_bal_dist}, Fig.~\ref{fig:auc_scores_pretrained}) and (ii) natural distribution splits that preserve the observed POPF prevalence (Table~\ref{tab:performance_comparison_3d_best_natural_dist}, Fig.~\ref{fig:auc_scores_trained_natural_dist}). In both settings, models operate on the same \emph{windowed} pancreas-centric CT inputs described in Sec.~\ref{sec:mask_pipeline}.

\subsubsection{Balanced splits} Under balanced evaluation, the residual spatiotemporal backbones consistently outperformed the CNN3D baseline, particularly when using MS masks. In the late arterial cohort, the best discrimination was achieved by R2Plus1DResNet18 and MC3ResNet18 with MS (AUC $\approx 0.73$), exceeding CNN3D (AUC $\approx 0.68$). A similar pattern held in the portal venous cohort, where R2Plus1DResNet18 with MS achieved the strongest performance (AUC $\approx 0.73$). Using TS masks reduced AUC across models (typically to $\approx 0.65$--$0.66$), indicating that ROI fidelity and segmentation quality strongly influence downstream prediction.

\subsubsection{Natural distribution splits} When preserving the natural class distribution (Table~\ref{tab:performance_comparison_3d_best_natural_dist}), discrimination remained strongest for the ResNet-based models, but the overall operating characteristics shifted as expected under class imbalance: precision decreased while recall remained comparatively high. With MS masks, MC3ResNet18 achieved the highest late arterial AUC (AUC $0.72\!\pm\!0.05$), followed closely by R2Plus1DResNet18 (AUC $0.68\!\pm\!0.07$); in portal venous scans, R2Plus1DResNet18 achieved the best AUC (AUC $0.70\!\pm\!0.03$). These results complement the balanced evaluation by reflecting a more realistic deployment scenario, where prevalence-driven shifts primarily impact precision/F1 while maintaining similar model ranking by AUC.

\section{Limitations}
This study has several limitations. First, although preoperative CT captures important anatomic and tissue characteristics relevant to POPF risk, imaging alone is unlikely to fully determine fistula development, which is also influenced by patient-level factors (e.g., age, BMI, sex, comorbidities) and operative/intraoperative variables (e.g., blood loss). Second, our pancreas-centric preprocessing and automated segmentation may introduce error propagation: imperfect masks, phase misclassification, or variability in CT acquisition and reconstruction can affect ROI definition and downstream predictions. Despite cross-validation, external validation on independent multi-institution cohorts and prospective evaluation are needed to confirm robustness, calibration, and clinical utility before deployment.

\section{CONCLUSION}
We developed a pancreas-centric deep learning pipeline to predict clinically relevant POPF from preoperative CT using automated pancreas segmentation, tight ROI cropping, and \emph{HU-windowed} intensity standardization prior to volume-based 3D classification. Across 5-fold cross-validation, residual spatiotemporal backbones (ResNet-(2+1)D-18 and ResNet-MC3-18) consistently outperformed a compact CNN3D baseline (AUC $\approx 0.73$ on balanced splits and AUC $\approx 0.70$--$0.72$ under natural prevalence). Together, these results support the feasibility of windowed-CT–only POPF risk stratification and motivate future work on improved segmentation/phase-aware modeling and external validation in independent cohorts. Overall, these results suggest that pancreas-centric, HU-windowed preoperative CT contains measurable imaging signatures associated with CR-POPF, enabling improved preoperative risk understanding and stratification.

\section{Compliance with Ethical Standards}
This study was approved by the Institutional Review Board of Mayo Clinic (IRB protocol 22-007521 and 24-011099) and conducted in accordance with applicable ethical guidelines and regulations.

\bibliographystyle{IEEEtran}
\bibliography{refs}

\end{document}